# RWESummary: A Framework and Test for Choosing Large Language Models to Summarize Real-World Evidence (RWE) Studies

Arjun Mukerji[1*], Michael L. Jackson[1], Jason Jones[1], Neil Sanghavi[1]
[1] Atropos Health; New York, NY, USA

*To whom correspondence should be addressed; arjun.mukerji@atroposhealth.com

## Abstract

Large Language Models (LLMs) have been extensively evaluated for general summarization tasks as well as medical research assistance, but they have not been specifically evaluated for the task of summarizing real-world evidence (RWE) from structured output of RWE studies. We introduce RWESummary, a proposed addition to the MedHELM framework (Bedi, Cui, Fuentes, Unell et al., 2025) to enable benchmarking of LLMs for this task. RWESummary includes one scenario and three evaluations covering major types of errors observed in summarization of medical research studies and was developed using Atropos Health proprietary data. Additionally, we use RWESummary to compare the performance of different LLMs in our internal RWE summarization tool. At the time of publication, with 13 distinct RWE studies, we found the Gemini 2.5 models performed best overall (both Flash and Pro). We suggest RWESummary as a novel and useful foundation model benchmark for real-world evidence study summarization.

## Background

Large databases linking electronic health records from many health care providers and payers (such as the Food and Drug Administration's Sentinel program (1) or the Observational Health Data Science Initiative (OHDSI)(2)) are enabling the generation of Real-World Evidence (RWE) both rapidly and at scale. An example is OHDSI's "How Often" project, which aims to estimate the incidence of hundreds of conditions across dozens of stratifications from many contributing data partners.(3) As RWE generation becomes more automated, a parallel need is for tools that can accurately summarize RWE results from structured study output. Large language models (LLMs) have shown promise for these types of summarization tasks.(4–6)

As with any use of LLMs, benchmarking is an essential step in evaluating whether an LLM can accurately summarize RWE results.(7) Several tools have been used to benchmark LLM performance for tasks in biomedicine, ranging from preexisting medical licensing exams (8) to custom-built clinical vignettes.(9) Perhaps the most comprehensive is MedHELM, a suite of tests covering diverse LLM use cases in medical practice.(10)

Current benchmarks do not explicitly evaluate the ability of LLMs to summarize structured RWE output. Building specific benchmarks for this use case is important, because LLMs may perform well in one task and yet struggle with seemingly related tasks.(11,12) In this manuscript, we describe RWESummary, a benchmark for testing how accurately LLMs can summarize structured RWE output. RWESummary currently consists of one scenario, an associated dataset, and three evaluations and is built to be a potential addition to the MedHELM suite of benchmarks.

## Methods

We implemented RWESummary as a new benchmark in the extensible HELM framework and (7) believe it would be appropriate for inclusion in the MedHELM suite.(10) The RWESummary benchmark currently consists of one scenario (i.e. use case), an associated dataset, and evaluation metrics. Each is described in detail below.

### Scenario

Atropos Health generates RWE studies in response to questions from clinicians and other clients. Each RWE study includes structured representations of the initial question, study design, and results. The LLM task is to write a plain English summary of the results while interpreting the results to answer the initial question. Note that this is different from other summarization use cases, where an LLM may be fed a long document (such as a scientific manuscript) and required to condense the paper to a short summary.(13) The scenario for this project assesses the ability of an LLM to extract, digest, and present appropriate information without the long context of a scientific paper.

### Dataset

We curated a new dataset of RWE studies for this task based on Atropos Health proprietary data. For this demonstration project the dataset consists of 13 individual RWE studies but is widely extensible to include more studies, additional therapeutic areas, and different study designs. The data are sourced from inputs to our internal RWE summarization tool and therefore cannot be made public. For each study, we extracted the original question, the study design in population-intervention-comparator-outcome-timeframe (PICOT) format, and results such as sample sizes, covariate distributions, and effect estimates with p-values and confidence intervals.

### Prompt

Our summarization prompt presents structured information in sections (Appendix 1). We begin with summarization instructions for our internal use case. The instructions differ slightly depending on the design of the study (cohort, case series, or cross-sectional). We then include the study question, design in PICOT format, and results. For cohort studies where baseline differences are corrected for by propensity score matching (14), significant baseline differences are included in the prompt. All results are included, even those not deemed significant. Finally, we include example summaries to help guide the model's output.

An example of the question, design, and results sections of the prompt can be found below along with an example summary.

```
Question: What is the impact of receipt of GLP-1 receptor agonists versus SGLT2i on
hospitalizations in patients with respiratory disease?

Study Design:
{
 "population": "Adults aged 18 years or older with a diagnosis of respiratory disease
(Asthma, COPD or bronchiectasis)
Minimum history of 90 days and minimum follow up of 3 years
",
 "intervention": "Receipt of GLP-1 receptor agonist within 1 year of diagnosis = index
on first receipt of diagnosis
No receipt of SGLT2i anytime ever"
 "control": "Receipt of SGLT2i within 1 year of diagnosis = index on first receipt of
diagnosis
No receipt of GLP-1 receptor agonist anytime ever"
 "timeframe": "2015-01-01 - 2024-12-27"
}

Results:
{
 "intervention_size": "34273",
 "control_size": "20571",
 "baseline_differences": {"female_pct": "control 47.3579 vs intervention 71.2047",
"age_mean": "control 64.2264 vs intervention 57.0941",
"comorb_congestive.heart.failure_pct": "control 26.4887 vs intervention 16.141",
"comorb_chronic.pulmonary.disease_pct": "control 86.3886 vs intervention 95.4279"}
 "outcomes": {'outcome_3yrs.ip.visit.for.asthma.exacerbation_bin!OR': '0.866 (0.665,
1.13)', 'outcome_3yrs.ip.visit.for.asthma.exacerbation_bin!pvalue': '0.28',
'outcome_3yrs.ip.visit.for.asthma_bin!OR': '1.03 (0.919, 1.14)',
'outcome_3yrs.ip.visit.for.asthma_bin!pvalue': '0.66',
'outcome_3yrs.ip.visit.for.bronchiectasis.exacerbation_bin!OR': '0.4 (0.155, 1.03)',
'outcome_3yrs.ip.visit.for.bronchiectasis.exacerbation_bin!pvalue': '0.058',
'outcome_3yrs.ip.visit.for.bronchiectasis_bin!OR': '0.624 (0.379, 1.03)',
'outcome_3yrs.ip.visit.for.bronchiectasis_bin!pvalue': '0.065',
'outcome_3yrs.ip.visit.for.copd.exacerbation_bin!OR': '0.988 (0.863, 1.13)',
'outcome_3yrs.ip.visit.for.copd.exacerbation_bin!pvalue': '0.86',
'outcome_3yrs.ip.visit.for.copd_bin!OR': '0.969 (0.889, 1.06)',
'outcome_3yrs.ip.visit.for.copd_bin!pvalue': '0.48',
'outcome_3yrs.ip.visit.for.gout_bin!OR': '0.858 (0.696, 1.06)',
'outcome_3yrs.ip.visit.for.gout_bin!pvalue': '0.151'}
}
```

Example LLM-generated summary:
This cohort study investigated the impact of GLP-1 receptor agonists (n=34,273) versus SGLT2i (n=20,571) on hospitalizations in adults with respiratory disease (asthma, COPD, or bronchiectasis) from 2015-2024. Baseline characteristics showed the GLP-1 group was younger (mean age 57.1 vs 64.2 years), had more females (71.2% vs 47.4%), more chronic pulmonary disease (95.4% vs 86.4%), but less congestive heart failure (16.1% vs 26.5%). After propensity score matching, no statistically significant differences were observed in 3-year inpatient visits for asthma exacerbation (p=0.28),

```
asthma (p=0.66), bronchiectasis exacerbation (p=0.058), bronchiectasis (p=0.065), COPD
exacerbation (p=0.86), COPD (p=0.48), or gout (p=0.151).
```

## Evaluations

We developed three separate evaluation metrics for RWESummary which cover the three most important classes of RWE summarization errors we have observed. All three require some level of semantic understanding and reasoning capability, so we rely on LLMs to grade responses rather than using purely mathematical measures like BERTscore.(15) Following Bedi et al (10) we used a three-member LLM jury for each metric, though with more recent models, namely OpenAI o4-mini, Anthropic Claude Sonnet 4, and Google Gemini 2.5 Pro.

With an LLM jury, each jury "member" rates each LLM-generated summary for the three outcome metrics described below. The final score for the summary is the mean of the jury scores. An important difference from Bedi et al (10) is that the LLM jury was tasked with making separate evaluations for each outcome metric rather than reporting all three metrics in a single evaluation. The rationale for this change was that pilot testing found that the LLM jury performed worse when asked to evaluate all outcomes in one pass compared to evaluating each outcome separately.

Our three primary outcome metrics were:
- Direction of effect. The LLM jury assessed whether the reported directions of estimated effects (positive, negative, neutral) matched the estimated effects in the RWE output.
- Accurate numbers. The LLM jury assessed whether the numbers reported in the summary correctly matched the numbers in the RWE output.
- Completeness. The LLM jury assessed whether all meaningful outcomes from the RWE output were included in the summary.

The prompts used for each outcome are provided in Appendix 2. As a secondary outcome we computed the total inference time used for each model.

## Models tested

Using our proposed scenario, we benchmarked the performance of nine distinct LLMs from three providers. Anthropic's Claude models (16) are integrated language and image models trained for generative output within a "constitutional AI" governance framework.(17) For this project we used Claude 3.7 Sonnet (2025-02-19), Claude 4 Sonnet (2025-05-14) and Claude 4 Opus (2025-05-14). Google's Gemini models are natively multimodal algorithms designed to interact with and generate text, audio, and images, among other data types.(18) For this project we used Gemini 2.0 Flash, Gemini 2.5 Flash, and Gemini 2.5 Pro. Finally, we used three models from OpenAI: o3 and o4-mini (both 2025-04-16 (19)) and GPT-4.1 (2025-04-14 (20)).

### Weighted scoring rubric

The evaluations described here consist of four separate outcome metrics. In some cases, users may wish to combine these metrics into a single scoring rubric to compare models. For illustrative purposes we demonstrate one potential approach to generating a weighted rubric. Weights for the different outcomes can be customized based on their perceived relative importance. We used the highest weight for the direction of effect outcome, reasoning that falsely reporting a protective drug effect as a harm (or vice versa) would be worse than inaccurate estimates or missing potentially important outcomes. After assigning weights, scores for each outcome are normalized based on the minimum and maximum observed scores for the outcomes. The weights are then applied to the normalized measures and the summed products are renormalized to generate a final score.

### Code Availability

The code for this paper is publicly available on GitHub. The RWESummary additions are in this commit.

## Results

All nine models were able to generate summaries for all 13 RWE studies. The models provided results that were generally judged to be complete (score range, 0.82 to 1.0), indicating that the important RWE outcomes tended to be included in the summary (Table 1). However, the models displayed more variability in correctly describing the direction of the reported effects (range, 0.72 to 0.95) and in the accuracy of reported numbers in the summaries (range, 0.77 to 0.97). Inference times varied widely (range, 2.2 to 22.8 seconds).

**Figure 1:** Outcome metrics for nine models summarizing RWE output. Blue shading indicates performance within 1% of the top-rated model in each category; red shading indicates performance within 1% of the bottom-rated model.

| Model | Direction of Effect | Numbers Accurate | Completeness | Inference Time |
|---|---|---|---|---|
| Gemini 2.5 Flash | 0.923 | 0.949 | **1** | 14.355 |
| Gemini 2.5 Pro | **0.949** | **0.974** | 0.897 | 14.226 |
| Gemini 2.0 Flash | 0.897 | **0.974** | 0.949 | **2.253** |
| o4-mini | 0.846 | 0.923 | **1** | 20.477 |
| o3 | 0.821 | 0.897 | **1** | 22.788 |
| Claude 4 Sonnet | 0.923 | 0.795 | 0.821 | 6.922 |
| GPT-4.1 | 0.795 | 0.897 | 0.949 | 4.456 |
| Claude 3.7 Sonnet | 0.718 | 0.821 | 0.949 | 5.735 |
| Claude 4 Opus | 0.744 | 0.769 | 0.949 | 9.585 |

No single model was the best or the worst performer across all outcome measures. Gemini 2.5 Pro was the top performer for effect direction and numeric accuracy. However, Gemini 2.5, o4-mini, and o3 were the top performers for completeness. Gemini 2.0 Flash was the fastest. Applying our example weighting scheme, the three Gemini models outperformed the other six models (Figure 2).

**Figure 2:** Application of example weights to model outcomes and computation of overall performance score

| Outcome weight | 20 | 6 | 5 | 3 | |
|---|---|---|---|---|---|
| Model | Direction of Effect | Numbers Accurate | Completeness | Inference Time | Score |
| Gemini 2.5 Flash | 89% | 88% | 100% | 41% | 99% |
| Gemini 2.5 Pro | 100% | 100% | 42% | 42% | 100% |
| Gemini 2.0 Flash | 77% | 100% | 72% | 100% | 94% |
| o4-mini | 55% | 75% | 100% | 11% | 61% |
| o3 | 45% | 62% | 100% | 0% | 46% |
| Claude 4 Sonnet | 89% | 13% | 0% | 77% | 61% |
| GPT-4.1 | 33% | 62% | 72% | 89% | 42% |
| Claude 3.7 Sonnet | 0% | 25% | 72% | 83% | 0% |
| Claude 4 Opus | 11% | 0% | 72% | 64% | 1% |
| *Normalizing Min* | *0.718* | *0.769* | *0.821* | *2.253* | |
| *Normalizing Max* | *0.949* | *0.974* | *1.000* | *22.788* | |

Other normalization and weighting mechanisms are of course possible. For example, holding everything else constant, if the weighting for inference time is set to 4 or 5, Gemini 2.0 Flash ties with the other Gemini models. If the inference time weighting is set to 6 or higher, Gemini 2.0 Flash becomes preferred. On the other hand, if we restore the inference time weight to 3 but change the normalizing max to "1.0" for direction of effect and numeric accuracy, Gemini 2.5 Flash becomes the sole "winner."

## Discussion

Advancements in computing power, EHR database linkages, and automated causal inference methods are changing the speed at which rigorous RWE can be generated. (21) A decade ago, generating RWE to answer a single clinical question would typically involve months to years of effort from a team of subject matter experts, data engineers, and statisticians. Today, machines can perform data engineering and analyses with minimal guidance. This shortens the process of ideation, iteration, and conclusion to days or hours. (22) LLMs that can accurately summarize the resulting RWE for the target audience will further improve the pace at which actionable RWE can be generated and disseminated. This paper presents a framework for evaluating the ability of LLM models to accurately generate such RWE summaries.

A number of studies have evaluated the ability of LLMs to extract numeric data from various sources and to structure those data in JSON or other machine-readable format (e.g. (23–25)). In biomedicine, studies have also evaluated the ability of LLMs to condense and summarize clinical data, potentially across multiple data sources. (5,26) These applications differ from the present use case, and few studies have explored the ability of LLMs to take tabular or structured research output and convert those to plain text summaries.

There are important limitations to this work. First, the study sample size of only 13 RWE test cases is small and should be increased and maintained over time. Both relative and absolute performance of the nine models tested should be viewed as provisional. Second, we used an LLM jury for testing. While LLM juries have performed well in other situations, (10,27) the particular jury used should be validated by human review in future applications. Third, as with any LLM use, our specific summarization results are dependent upon the prompts, which have not yet been optimized for each model. Fourth, results depend both upon the substance and structure (prompts) of RWE inputs. We elected to use for this study what we use in practice to facilitate judging whether the results made sense and were useful.

The spaces of RWE and LLMs are context dependent and evolving rapidly. We do not believe there is likely to be a model to rule them all and for all time in the short to mid-term. We must make pragmatic decisions to choose how to automate RWE summarization in a narrative form in order to take advantage of the other technologies facilitating the generation and provisioning of RWE at the bedside and for policy decisions. We hope the methods we have laid out facilitate driving pragmatic decisions around LLM model selection with data as we aspire to do the same via RWE in the broader healthcare context.

# APPENDIX 1: Summarization prompt

Generate a scientific summary for a cohort study in about 100 words like the example summary delimited by ------ for the question below. Make sure baseline differences relevant to the question are mentioned in the summary, but don't claim they are significant. Also make sure all outcomes with a p-value of less than 0.05 are reported as significant in the summary, while outcomes with a p-value of 0.05 or greater are reported as non-signficant. Note that these results are after propensity score matching. If no outcomes are provided, make no statements about results or significance. Report all percentages and Charlson scores to one decimal place.

The study design and results are provided in JSON format below.

Question: What is the impact of receipt of GLP-1 receptor agonists versus SGLT2i on hospitalizations in patients with respiratory disease?

Study Design:
{
 "population": "Adults aged 18 years or older with a diagnosis of respiratory disease (Asthma, COPD or bronchiectasis)
Minimum history of 90 days and minimum follow up of 3 years
",
 "intervention": "Receipt of GLP-1 receptor agonist within 1 year of diagnosis = index on first receipt of diagnosis
No receipt of SGLT2i anytime ever"
 "control": "Receipt of SGLT2i within 1 year of diagnosis = index on first receipt of diagnosis
No receipt of GLP-1 receptor agonist anytime ever"
 "timeframe": "2015-01-01 - 2024-12-27"
}

Results:
{
 "intervention_size": "34273",
 "control_size": "20571",
 "baseline_differences": {"female_pct": "control 47.3579 vs intervention 71.2047", "age_mean": "control 64.2264 vs intervention 57.0941", "comorb_congestive.heart.failure_pct": "control 26.4887 vs intervention 16.141", "comorb_chronic.pulmonary.disease_pct": "control 86.3886 vs intervention 95.4279"}
 "outcomes": {'outcome_3yrs.ip.visit.for.asthma.exacerbation_bin!OR': '0.866 (0.665, 1.13)',
'outcome_3yrs.ip.visit.for.asthma.exacerbation_bin!pvalue': '0.28',
'outcome_3yrs.ip.visit.for.asthma_bin!OR': '1.03 (0.919, 1.14)',
'outcome_3yrs.ip.visit.for.asthma_bin!pvalue': '0.66',
'outcome_3yrs.ip.visit.for.bronchiectasis.exacerbation_bin!OR': '0.4 (0.155, 1.03)',
'outcome_3yrs.ip.visit.for.bronchiectasis.exacerbation_bin!pvalue': '0.058',

```
'outcome_3yrs.ip.visit.for.bronchiectasis_bin!OR': '0.624 (0.379, 1.03)',
'outcome_3yrs.ip.visit.for.bronchiectasis_bin!pvalue': '0.065',
'outcome_3yrs.ip.visit.for.copd.exacerbation_bin!OR': '0.988 (0.863, 1.13)',
'outcome_3yrs.ip.visit.for.copd.exacerbation_bin!pvalue': '0.86',
'outcome_3yrs.ip.visit.for.copd_bin!OR': '0.969 (0.889, 1.06)',
'outcome_3yrs.ip.visit.for.copd_bin!pvalue': '0.48', 'outcome_3yrs.ip.visit.for.gout_bin!OR': '0.858 (0.696, 1.06)', 'outcome_3yrs.ip.visit.for.gout_bin!pvalue': '0.151'}
}
```
---------------------------------

Summary example 1: We identified patients within a national dataset who had IBD that either received anti-TNF therapy (n=4,224) or never received biologic while still receiving mesalamine/sulfasalazine therapy (n=8,630) from 2010-present.
The intervention and control were similar in sex, race/ethnicity, and baseline comorbidity scores (2.1 vs 1.2).
However, the cohort receiving anti-TNF was significantly younger (56.5 years vs 47 years).
The patients were evaluated for specific GI infections as well as composite infections over 5-years using high-dimensionality propensity score matching.
The cohort receiving anti-TNF had higher rates of C-Diff as well as Bacterial Intestinal infection and lower rates of non-bacterial intestinal infections.

Summary example 2: In this cohort study, we evaluated the risk of aspiration pneumonia in obese or diabetic patients undergoing colonoscopy, comparing those who used GLP-1 agonists (n=1,478) within 90 days prior to the procedure against those who did not use GLP-1 agonists but were on other medications (n=26,189) from 2015 to the present.
The study excluded patients with conditions like achalasia, esophagus dyskinesia, and those who had been mechanically ventilated close to the procedure.
Notably, the intervention group had a higher percentage of patients with diabetes and its complications, chronic pulmonary disease, mild liver disease, and renal disease compared to the control group.
Despite these baseline differences, the outcomes showed an odds ratio of 0.5 for aspiration pneumonia in the GLP-1 agonist group, with a wide confidence interval (0.00847, 9.61) and a p-value of 1, indicating no statistically significant difference in the risk of aspiration pneumonia between the two groups.

# APPENDIX 2: Evaluation Prompts

DIRECTION_PROMPT_TEMPLATE = """# Instructions
You are an expert in medical research.
I will give you 2 inputs.
The first is a prompt to summarize a retrospective observational research study.
The second is a summary generated from the prompt.
Your job is to judge whether the direction of all effects reported in the summary matches the prompt.

## Specific rules
- Treat all continuous outcomes (cont) with a positive value as an increase over baseline.
- All mean differences (MD) are Intervention minus Control.
- Only evaluate if the direction of the effect (positive/negative, increase/decrease) reported in the summary matches the prompt
 - However, if an effect is not significant, it is acceptable to report as "no difference" even if the prompt numerically shows an increase or decrease.

# Response
Your response should be either 0 or 1. 1 indicates the direction of all effects is reported correctly. Otherwise, respond with 0. Only mention one of these two responses.

# Inputs
[BEGIN DATA]
The summarization prompt is: {{QUESTION}}.
The summary is: {{RESPONSE}}.
[END DATA]

# Output Format
Generate a valid JSON object with your evaluation:
{
  "direction_of_effect": {
    "score": 0,
    "explanation": "Explain why this score was given."
  },
}

Ensure the output is valid JSON:
- Use **double quotes** (") for all keys and string values.
- When quoting text or sections inside the explanations, use escaped double quotes (\") to
 maintain valid JSON formatting.
- Do not include any additional information in the output.
"""

NUMBERS_PROMPT_TEMPLATE = """# Instructions
You are an expert in medical research.
I will give you 2 inputs.
The first is a prompt to summarize a retrospective observational research study.
The second is a summary generated from the prompt.
Your job is to judge whether all numbers in the summary are accurately sourced from the prompt with reasonable rounding.

## Specific rules
- Only evaluate if the numbers in the summary match the relevant source in the prompt.
- Rounding of decimals is allowed, so long as it is numerically accurate.

# Response
Your response should be either 0 or 1. 1 indicates the numbers are all reported correctly. Otherwise, respond with 0. Only mention one of these two responses.

# Inputs
[BEGIN DATA]
The summarization prompt is: {{QUESTION}}.
The summary is: {{RESPONSE}}.
[END DATA]

# Output Format
Generate a valid JSON object with your evaluation:
{
  "numbers_accurate": {
    "score": 0,
    "explanation": "Explain why this score was given."
  },
}

Ensure the output is valid JSON:
- Use **double quotes** (") for all keys and string values.
- When quoting text or sections inside the explanations, use escaped double quotes (\") to maintain valid JSON formatting.
- Do not include any additional information in the output.
"""

COMPLETENESS_PROMPT_TEMPLATE = """# Instructions

You are an expert in medical research.
I will give you 2 inputs.
The first is a prompt to summarize a retrospective observational research study.
The second is a summary generated from the prompt.
Your job is to judge whether all significant outcomes in the prompt are mentioned in the summary.

## Specific rules
- Check if the prompt contains an outcome ending in '!pvalue' in the "outcomes" dictionary.
  - If so, and this p-value is <= 0.05, ensure the summary explicitly mentions that there is a difference or higher/lower risk with that outcome.
  - Note that not all studies have outcomes with p-values. In such cases, return "CORRECT".
- Non-significant outcomes do not need to be mentioned in the summary.

# Response
Your response should be either 0 or 1. 1 indicates all significant outcomes from the prompt are mentioned in the summary. Otherwise, respond with 0. Only mention one of these two responses.

# Inputs
[BEGIN DATA]
The summarization prompt is: {{QUESTION}}.
The summary is: {{RESPONSE}}.
[END DATA]

# Output Format
Generate a valid JSON object with your evaluation:
{
  "completeness": {
    "score": 0,
    "explanation": "Explain why this score was given."
  },
}

Ensure the output is valid JSON:
- Use **double quotes** (") for all keys and string values.
- When quoting text or sections inside the explanations, use escaped double quotes (\") to maintain valid JSON formatting.
- Do not include any additional information in the output.
"""